\pdfoutput=1
\documentclass[11pt]{article}
\DeclareUnicodeCharacter{01BF}{w}
\DeclareUnicodeCharacter{01F7}{W}
\DeclareUnicodeCharacter{021D}{g}

\usepackage{acl}

\usepackage{times}
\usepackage{latexsym}

\usepackage[T1]{fontenc}
\usepackage[utf8]{inputenc}

\usepackage{microtype}

\usepackage{inconsolata}

\usepackage{graphicx}
\usepackage{linguex}

\usepackage{booktabs}
\usepackage{tikz-dependency}

\title{Building UD Cairo for Old English in the Classroom}

\author{Lauren Levine, Junghyun Min, and Amir Zeldes \\
        Georgetown University, Department of Linguistics\\
        \texttt{\{lel76, jm3743, amir.zeldes\}@georgetown.edu}
}

\begin{document}
\maketitle
\begin{abstract}
In this paper we present a sample treebank for Old English based on the UD Cairo sentences, collected and annotated as part of a classroom curriculum in Historical Linguistics. To collect the data, a sample of 20 sentences illustrating a range of syntactic constructions in the world's languages, we employ a combination of LLM prompting and searches in authentic Old English data. For annotation we assigned sentences to multiple students with limited prior exposure to UD, whose annotations we compare and adjudicate. Our results suggest that while current LLM outputs in Old English do not reflect authentic syntax, this can be mitigated by post-editing, and that although beginner annotators do not possess enough background to complete the task perfectly, taken together they can produce good results and learn from the experience. We also conduct preliminary parsing experiments using Modern English training data, and find that although performance on Old English is poor, parsing on annotated features (lemma, hyperlemma, gloss) leads to improved performance.

\end{abstract}

\section{Introduction}

Treebanking historical languages is a challenging task for multiple reasons -- to name just a few: 1. annotators are never native speakers; 2. it is difficult to recruit annotators with sufficient background in the language, in theoretical syntax, and in corpus annotation; 3. guidelines and sample datasets illustrating annotation principles are often tailored to modern languages, with examples that would be implausible in context for ancient texts; 4.  there is a scarcity of previous work for annotators to reliably refer to, making examples for contexts that are common in ancient texts difficult to find.

At the same time, tackling ancient language syntax through treebanking can be a rewarding and pedagogically valuable way of confronting students of ancient languages and historical linguistics with authentic syntax in ways that simply looking at an original sentence and its translation cannot \cite{bamman-crane-2007-latin,Mambrini2016,keersmaekers-etal-2019-creating}. An additional challenge here is defining a realistic target for treebanking which can be completed as a homework assignment without becoming overwhelming.

In Universal Dependencies (UD, \citealt{de-marneffe-etal-2021-universal}), the most popular treebanking formalism, there is no single template for starting a treebanking project. However, a set of basic sentences is sometimes suggested, known as the Cairo sentences and originally proposed at CICLing 2015 in Cairo.\footnote{\url{https://github.com/UniversalDependencies/cairo}} The data has been translated and used as a UD starting point for a number of languages, including Ligurian \cite{lusito-maillard-2021-universal}, Luxembourgish \cite{plum-etal-2024-luxbank} and Gujarati \cite{jobanputra-etal-2024-universal}. In this paper, we therefore propose and report on a classroom historical treebanking activity creating a UD Cairo-style corpus for Old English (Ald-Ænglisc or OE), including generating the text, treebanking proper, and adding supplementary annotations. 

Although some work on treebanking OE and later stages of older English exists -- notably The York–Toronto–Helsinki Parsed Corpus of Old English Prose \citep[YCOE; ][]{Taylor2007York}, the Penn-Helsinki Parsed Corpus of Early Modern English \citep[PPCEME; ][]{kroch2004penn}, and their follow-up work \citep{kroch2020penn, kulick-etal-2023-parsing} -- no such data exists in UD. We add to the literature by providing the first OE contribution to UD. We evaluate treebanking from students with and without previous exposure to UD, and run some parsing experiments to reveal how similar or distinct OE syntax is compared to Modern English (ME). We release all data publicly\footnote{\url{https://universaldependencies.org/treebanks/ang_cairo/index.html}} and hope this work inspires others to incorporate more treebanking into classroom activities.

\section{LLM-assisted UD Cairo in Ald-Ænglisc}

Translating the UD Cairo sentences to Old English is a non-trivial task, especially considering the goal of expressing the basic constructions of the language in the most natural possible way, without introducing translation effects conforming to the Modern English syntax or lexicon. As a first step, we considered simply asking an LLM, in this case GPT4o, to translate the sentences (see Appendix \ref{sec:appendix-gpt} for the full list of sentences with initial GPT outputs and final versions).

Inspecting GPT's output revealed that while the model was surprisingly capable of translating the Cairo sentences into superficially correct Old English, many problems emerged, especially in morphology and the tendency to mimic the words and word order of the input one for one. For example, consider sentence \ref{ex:letter1}:

\ex. \label{ex:letter1}
 \a. The girl wrote a letter to her friend 
 \b. Sēo mǣden wrāt ǣnne ærendgecwidd tō hyre frēonde.

The model translated `the girl' into `seo mæden' - a correct lexical choice with an incorrect article gender (feminine, when `mæden' was in fact neuter in OE). The model also tended to insert indefinite articles (here `ǣnne'), which were not yet grammaticalized or expected in OE, where the Modern English original had them, resulting in superfluous `ones' (e.g.~`one letter' for `a letter', where just `letter' would be correct). We also note that GPT recreates the input word order with the `to' dative, rather than using the more usual ditransitive construction, (`wrote her friend a letter'), which was used more in OE. Finally, the word for `letter' is a creative invention, compounded from two attested OE stems, but which is not attested as a word. In the final data, we therefore corrected the gender error, removed the indefinite article, used the ditransitive word order, and used Bosworth Toller's Anglo-Saxon Online Dictionary\footnote{\url{https://bosworthtoller.com/}} to select an attested word for `letter', resulting in \ref{ex:letter2}:

\ex. Þæt mæden wrat hyre freonde ærend-writ\label{ex:letter2}

At the same time, GPT was adept at proposing creative solutions to vocabulary gaps: for example, Cairo sentence 6 refers to washing a car, to which GPT, unprompted, added: ``(Note: "car" is anachronistic, but Old English did have "cār," meaning "chariot" or "cart.")'', and selected OE \textit{cræt} `cart' for its own translation, which we adopted. In some cases, reprompting for additional translation ideas was also an effective strategy.

%\begin{figure*}[h!bt]
%\centering
%\begin{dependency}[arc edge, arc angle=80, text only label, label style={above}]
%\begin{deptext}[column sep=.7cm]
%f \& na \& \v{c}oo \& s \& \v{c}e \& ou \& rmnkah \& an \& pe  \\
%3SGM \& FUT \& say \& 3SGF \& that \& a \& earth-man \& not \& COP \\
%Root=x \& FUT \& say \& 3SGF \& that \& a \& earth-man \& not \& COP \\
%\end{deptext}
%\deproot{3}{root}
%\depedge{3}{1}{nsubj}
%\depedge{3}{2}{aux}
%\depedge{3}{4}{obj}
%\depedge{7}{5}{mark}
%\depedge{7}{6}{det}
%\depedge{7}{8}{advmod}
%\depedge{7}{9}{cop}
%\depedge{3}{7}{ccomp}
%\end{dependency}
%\caption{Analysis of a doubled object clause construction: \textit{He would say (it) that he is not an earthly man}.}\label{joosje}
%\end{figure*}

As a final source to ensure our chosen constructions corresponded to attested Old English syntax, we cross-referenced sentences containing target constructions in a Modern English New Testament translation with the OE Wessex Gospels, the oldest existing manuscript of the Gospels in OE. For example, GPT's initial version of Cairo sentence 17 contains a counterfactual `she should have been doing', which GPT rendered with \textit{heo scolde þæt ... dyde}, roughly corresponding to a Modern English `she should that did'. However searching the gospels for `should have', we quickly realized the most common equivalent in the Wessex Gospels is actually \textit{hit gebyrede þæt heo dyde}, i.e.~``it was appropriate that she did...''. 

In summary, our approach to producing an ancient language version of the Cairo sentences is: 1. ask an LLM to translate the sentence; 2. check the result for morphological errors; 3. ensure chosen words are attested in a dictionary and replace/reprompt for alternatives as needed; 4. check a modern translation for equivalent modern construction instances, then ensure their original counterparts correspond to the selected constructions.

\section{Annotation}

\subsection{UD trees and classroom annotation}

Our primary objective was to design a classroom learning activity for an audience that might not be familiar with UD guidelines, rather than to create data for OE parsing. As part of a mid-level, mixed over/undergraduate course on Historical Linguistics, we formulated an assignment in the Old English module of the class in which students were asked to treebank four random OE Cairo sentences. We allowed students to access both UD corpora (using the Grew-match interface, \citealt{guillaume-2021-graph}) and Modern English parsers -- we pointed them to online interfaces for UDPipe \cite{straka-2018-udpipe} and Stanza \cite{qi-etal-2020-stanza}, but any parser was allowed. 

Since the students mostly had only minimal instruction about the UD guidelines, they were instructed to look in existing UD data for similar constructions to the sentence they were annotating, and to play with the parsers by feeding them what they felt were modern English equivalents (the original Cairo sentences in English were not provided, though students may have found them online too). Use of AI agents was also allowed, though we used the Arborator graphical interface \cite{guibon-etal-2020-collaborative}, meaning students could not simply let LLMs solve the assignment for them directly.

\begin{figure}[h!tb]
\centering
\includegraphics[width=0.5\textwidth]{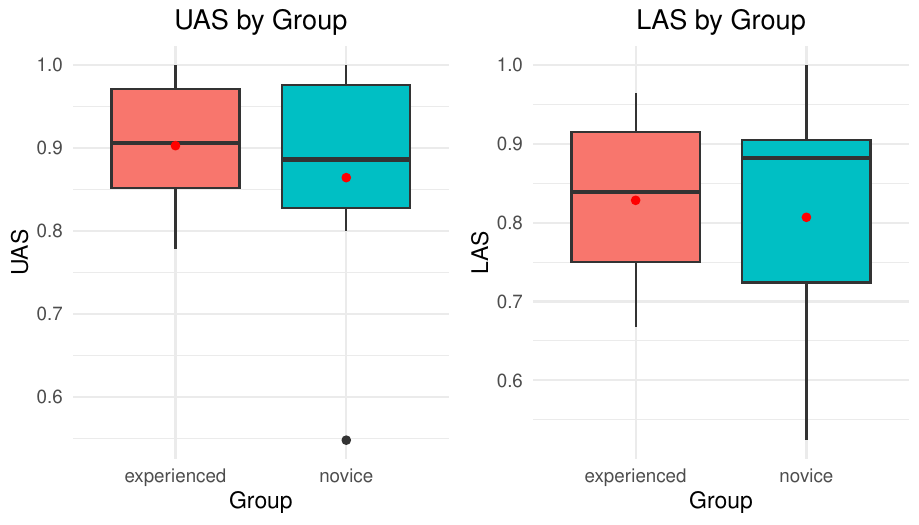}
\caption{Unlabeled and labeled attachment scores (UAS \& LAS) for experienced and novice annotators.}
\label{fig:student-scores}
\end{figure}

Figure \ref{fig:student-scores} plots students' accuracy on the assignment in terms of unlabeled and labeled attachment scores (UAS and LAS). We note that some of the students in the class, who had background in Computational Linguistics and previous exposure to UD annotation, unsurprisingly did better on average in terms of accuracy, and therefore we report scores separately for those two annotator subsets.

As Figure \ref{fig:student-scores} shows, while the mean scores (indicated with red dots) are somewhat higher for the experienced group, novice students were able to achieve rather comparable scores by using corpus searches and automatic modern English parsers as references, and the median novice LAS score is even higher than the experienced one. At the same time, the range of scores is wider for novices, which we suspect is due to the heterogeneity of their background in general linguistics, with students who had more background in theoretical (but non-UD) syntax achieving better accuracy.

We report inter-annotator agreement (IAA) metrics for our gold reference annotations (accuracy, Cohen's Kappa \citep{cohen-1960-kappa}, and F1-score) between two groups of experienced Modern English UD annotators in Table \ref{tab:iaa_results}. We also include IAA accuracy on section 23 of the Penn Treebank \citep{marcus-etal-1993-ptb} measured by \citet{berzak-etal-2016-anchoring} as a point of reference. As our numbers are similar to Penn Treebank ones, we believe experience with ME UD annotation translates non-trivially to performance on OE UD annotation.

\begin{table}[h!tb]
    \centering
    \begin{tabular}{lcc}
    \toprule
         \textbf{IAA score} & \textbf{UAS} & \textbf{LAS}  \\
         \midrule
         \textbf{UD Cairo OE} \\
           acc. & 94.15 & 88.30 \\
           $\kappa$ & 93.38 & 88.27  \\
           F1 & 96.99 & 93.79 \\
         \textbf{Reference} \\
           PTB-WSJ (acc.)
         & 93.07 & 88.32 \\
        \bottomrule
    \end{tabular}
\caption{Inter-annotator agreement metrics: accuracy, Cohen's Kappa, and F1-Score. We present an IAA metric on PTB-WSJ by \citet{berzak-etal-2016-anchoring} for reference.}
\label{tab:iaa_results}
\end{table}

% \citep{marcus-etal-1993-ptb}

\subsection{Disagreement Analysis}
\label{sec:disagreement}

\begin{figure}
\centering
\includegraphics[width=0.5\textwidth]{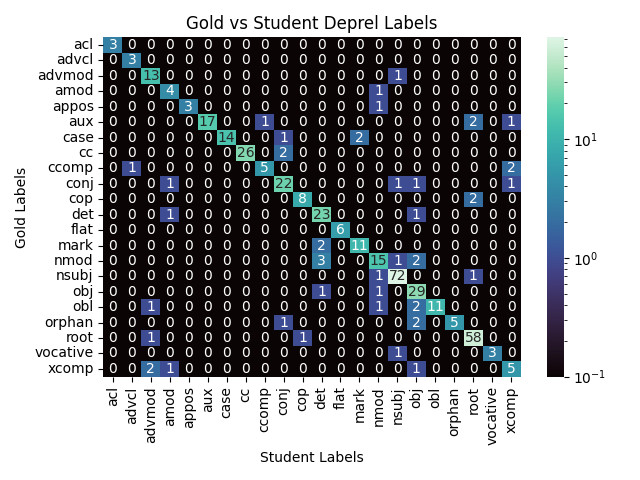}
\caption{Confusion matrix for gold dependency relation (deprel) labels vs the student annotations. Deprel subtypes are collapsed, only deprel labels that occur > 5 times are included, and \texttt{punct} was excluded.}
\label{fig:deprel_cm}
\end{figure}

In Figure \ref{fig:deprel_cm}, we show a confusion matrix comparing our final gold labels for the Cairo UD sentences with the student annotations from our classroom activity. In line with our high inter-annotator agreement scores in the previous section, we see that there is strong overlap in the selection of dependency relation labels. However, there is still a smattering of disagreements between the gold labels and the student labels. The dependency relation that was mislabeled by students the most was \texttt{nmod}, which was mislabeled 6 times (3 times as \texttt{det}, 2 times as \texttt{obj}, and 1 time as \texttt{nsubj}). These confusions primarily stemmed from students not understanding when to apply the \texttt{nmod:poss} subtype when possessive case was involved in the sentence. The label that was most misapplied by students was \texttt{obj}, which was misapplied a total of 9 times (2 times each on \texttt{nmod}, \texttt{obl}, and \texttt{orphan} relations, and 1 time each as on \texttt{conj}, \texttt{det}, and \texttt{xcomp} relations). In Figure \ref{fig:disagreement}, we show a dependency analysis for one of the UD Cairo sentences where a student both misapplied the \texttt{obj} label rather than utilizing the \texttt{orphan} relation, as well as incorrectly using \texttt{det} for a possessive pronoun. We see that the student uses a promotion strategy, rather than utilizing \texttt{orphan}. Likely the use of more complicated relations, such as \texttt{orphan}, could be learned sufficiently with more time devoted to annotator training. 

\begin{figure*}
\centering
\begin{dependency}[arc edge, arc angle=80, text only label, label style={above}]
\begin{deptext}[column sep=.7cm]
He \& bohte \& cræt \& ac \& his \& broþor \& hƿeol-bearƿe \\
He \& bought \& car \& but \& his \& brother \& wheelbarrow \\
\end{deptext}
\depedge{2}{1}{nsubj}
\depedge{2}{3}{obj}
\depedge{6}{4}{cc}
\depedge{2}{6}{conj}
\depedge{6}{5}{nmod:poss}
\depedge{6}{7}{orphan} 
\depedge[edge below, edge style={blue}, label style={text=blue}]{2}{1}{nsubj}
\depedge[edge below, edge style={blue}, label style={text=blue}]{2}{3}{obj}
\depedge[edge below, edge style={blue}, label style={text=blue}]{7}{4}{cc}
\depedge[edge below, edge style={blue}, label style={text=blue}]{6}{5}{det}
\depedge[edge below, edge style={blue}, label style={text=blue}]{7}{6}{nsubj}
\depedge[edge below, edge style={blue}, label style={text=blue}]{2}{7}{conj}
\depedge[edge below, edge style={blue}, label style={text=blue}]{7}{4}{cc}
\end{dependency}
\vspace{-6pt}
\caption{Dependency analysis of a ellipsis construction in OE: \textit{He bought a car but his brother just a wheelbarrow.} ME gloss is included below the OE tokens. The gold analysis is shown on top in black, and a student analysis is shown on the bottom in blue.}\label{fig:disagreement}
\vspace{-6pt}
\end{figure*}

\subsection{Additional annotations}
\label{sec:annotations}
In addition to the classroom UD annotations that comprise part of speech tags and dependency labels and their attachments, we provide morphology, hyperlemma, and root annotations (a sample analysis is included in Appendix \ref{appendix:sample}).

\textbf{Morphological features:} We follow UD's standard CoNLL-U format and provide annotations on case, number, gender, tense, person, verb-form, and mood. In our corpus, we count four cases (nominative, accusative, genitive, and dative), two numbers (singular and plural), three genders (masculine, neuter, and feminine), two tenses (past and present), three persons (1st, 2nd, and 3rd), three verb-forms (finite, infinitive, and participle), and two moods (indicative and imperative). However, the above list is not exhaustive for Old English morphology. In submitting our corpus to UD, we define Old English's language-specific morphological feature space as outlined in Appendix \ref{appendix:feats}.

\textbf{Gloss:} We provide word-level glosses in Modern English, whose uninflected base forms are often equivalent to the hyperlemma described below.

\textbf{Hyperlemmas:} Hyperlemmas, first proposed in \citet{dipper2004german}, align word forms and lemmas from different language branches and historical periods that correspond to each other. For example, a hyperlemma \textit{have} would normalize German \textit{haben}, English \textit{have}, Old English \textit{habban} and their inflected forms. We follow several annotated corpora of historic European languages \citep{meyer2011russian, kucera2014czech, shokina2016churchslavonic, chiarcos2018mhgerman} and include hyperlemmas in our annotations.

\textbf{Roots and language of origin:} Roots are another method of normalizing within language family and across historical periods. For example, German \textit{trinken}, English \textit{drink}, and Old English \textit{drincan} can all be associated with their reconstructed Indo-European root \textit{*d\textsuperscript{h}renǵ}. While it is more common for Afro-Asiatic corpora to contain root information \cite[e.g.][]{hajic-2009-prague}, we provide root information in addition to hyperlemmas. If the root is of non-Indo-European origin, its language of origin is also specified.

% could add specific annotation decision -> how we handled compounds, 

\section{Parsing experiments}

\begin{table}[]
\centering
\resizebox{\columnwidth}{!}{%
\begin{tabular}{lcccc}
\toprule
\multicolumn{1}{c}{} & \multicolumn{2}{c}{\textbf{ELECTRA} } & \multicolumn{2}{c}{\textbf{mBERT}} \\
\midrule
Input & UAS & LAS & UAS & LAS \\ \midrule
GUM 11 dev & 95.35 &  93.17 & 94.21 & 91.95 \\
Form & 23.98 & 12.87 & 41.52 & 25.73 \\
\hspace{2mm} +Norm & 35.09 & 20.47 & 41.52 & 27.49 \\
Lemma & 18.13 & 7.60 & 38.60 & 22.81 \\
\hspace{2mm} +Norm & 32.16 & 16.96 & 39.18 & 23.39 \\
Hyperlemma & 61.99 & 49.12 & 70.76 & 56.73 \\
\hspace{2mm} +Norm & 74.27 & 59.06 & 83.04 & 67.84 \\
Gloss & 81.87 & 75.44 & 85.38 & 78.36 \\ \bottomrule
\end{tabular}%
}
\caption{Unlabeled and labeled attachment scores (UAS \& LAS) for parsers trained on ME data (GUM V11).}
\vspace{-6pt}
\label{tab:parse_results}
\end{table}

%\multicolumn{1}{c}{} & \multicolumn{2}{c}{ \citep{clark2020electra}} & \multicolumn{2}{c}{ \citep{devlin-etal-2019-bert}} \\

In order to shed light on how much Old English dependency syntax analysis differs from that of Modern English, we conduct parsing experiments on our OE Cairo sentences using dependency parsers trained on ME. While it may be possible and potentially effective to use dependency parsers trained on other modern and historical Germanic languages, we limit the scope our parsing experiments to English. We train DiaParser \citep{dozat2017deep}, a neural biaffine dependency parser, using English ELECTRA \citep{clark2020electra} and multilingual mBERT \citep{devlin-etal-2019-bert} embeddings, in order to compare the impact of  monolingual vs multilingual pre-trained inputs. For training data, we use version 11 of the GUM corpus \cite{Zeldes2017}, which is the largest available UD corpus of Modern English, containing 268k tokens across a variety of genres. We evaluate the ELECTRA and mBERT models on our OE Cairo sentences, using our various token annotations (Form, Lemma, Hyperlemma, and Gloss) as input. The results of these experiments are reported in Table \ref{tab:parse_results}, giving the UAS and the LAS for each model. Form refers to the unaltered text of the tokens, and Lemma refers to the uninflected form of the tokens. Hyperlemma and Gloss are as described in Section \ref{sec:annotations}. $+$Norm indicates the addition of orthographic normalization to the annotation of the preceding row -- replacing OE letters that no longer exist in ME ("th" for "þ", "e" for "æ", etc.).

Looking at Table \ref{tab:parse_results}, we see that although the ELECTRA model has higher performance on the Modern English GUM 11 dev data, mBERT uniformly performs better on the Old English data for all annotation categories. This demonstrates the advantage that multilingual pretrained embeddings provide when a parsing model is used on a different language than the one used during model training. We also note that for the categories of Form and Lemma, orthographic normalization leads to large improvements for the ELECTRA model with monolingual embeddings, but only minor improvement to the mBERT model with multilingual embeddings. However, both models show a large increase in LAS between Hyperlemma and Hyperlemma+Norm, with an increase of 9.94\% for the ELECTRA model and an increase of 11.11\% for the mBERT model.

For both models, performance increases from Form to Hyperlemma and again to Gloss, for both UAS and LAS, with the mBERT model giving the highest score of UAS 85.38\% and LAS 78.36\%. Using Hyperlemma and Gloss for the model input allows us to abstract away from the lexical differences between Old English and Modern English. Although parsing the Gloss leads to a large improvement in performance over parsing the Form (LAS $\Delta$ 62.65\% for the ELECTRA model and LAS $\Delta$ 52.63\% for the mBERT model), there is still a large drop in performance when compared to the ME dev data (LAS $\Delta$ 17.73\% for ELECTRA and LAS $\Delta$ 13.59\% for mBERT). Although some degradation can be expected in cross-corpus parsing \cite{zeldes-schneider-2023-ud}, its magnitude indicates that there are notable syntactic differences between OE and ME for dependency parsing, even after controlling for lexical differences. Still, the scores for parsing on glossed tokens of Old English are high enough to suggest that leveraging Modern English models to parse glossed Old English texts could be an effective method to bootstrap the creation of Old English syntactic dependency datasets in the future.

\section{Conclusion}

In this paper, we presented an Old English version of the UD Cairo sentences, annotated for syntactic dependencies, morphological features, and historically relevant features of hyperlemma and root. In using LLMs for the initial generation of the OE version of the data, we demonstrate how current AI tools can be used in conjunction with quality assurance procedures in the workflow of data creation to produce more reliable results. By including the treebanking task as part of a course in Historical Linguistics, we show that a brief introduction to UD in the classroom, along with providing reference resources, can result in relatively good performance, even for a language with which most of the students have no familiarity. Additionally, through preliminary parsing experiments, we demonstrate that dependency parsers trained on Modern English data may be of use for the creation of new syntactic dependency datasets for Old English. We hope that the sample Old English treebank presented here can serve as a guide for future classroom annotation efforts.

%for OE and beyond.

\section*{Acknowledgments}

We would like to thank the students in the course Corpus Approaches to Historical Linguistics at Georgetown University for participating in the course and the creation of the dataset. The final public repository will include all annotation versions of all students who agreed to release their work and signed an open license agreement. Credit is given by name to all students who wished to be named, as well as anonymously to students who wish to remain anonymous.

\bibliography{custom}

\appendix

\begin{figure*}[h!tb]
\centering
\begin{dependency}[arc edge, arc angle=80, text only label, label style={above}]
\begin{deptext}[column sep=.7cm]
Þæt \& mæden \& wrat \& hyre \& freonde \& ærend-writ \& . \\ % ƿ -> w
the \& maiden \& write \& she \& friend \& errand-writ \& . \\
*so \& *mog$^h$ \& *wrey \& *ḱey \& *preyH \& *h$_2$eyeri-wrey \& \_ \\
Sing-Neut \& Sing-Neut \& 3-Sing \& Sing-Fem \& Sing-Masc \& Sing-Neut \& \_ \\
Nom \& Nom \& Fin-Ind-Pret \& Gen \& Dat \& Acc \& \_ \\
\end{deptext}
\deproot{3}{root}
\depedge{2}{1}{det}
\depedge{3}{2}{nsubj}
\depedge{5}{4}{nmod:poss}
\depedge{3}{5}{iobj}
\depedge{3}{6}{obj}
\depedge{3}{7}{punct}
\end{dependency}
\caption{Dependency analysis of a ditransitive construction in OE: \textit{The girl wrote a letter to her friend}. The glossing contains (from top to bottom) OE form, hyperlemma, root, and morphological feature annotations (person-number-gender and verbForm-mood-tense, as applicable).}\label{fig:sample}
\end{figure*}
%  (person-number-gender-case-tense, as applicable)
% SG-N-Nom \& SG-N-Nom \& 3-SG-Pret-Fin-Ind \& SG-F-Gen \& SG-M-Dat \& SG-N-Acc \& \_ \\

\section{GPT initial outputs}
\label{sec:appendix-gpt}

GPT's initial renditions of the Old English Cairo sentences are given below, in italics. Note that these include clear errors. For reference, original English sentences from the Cairo dataset in quotes and the final Old English translations for our UD Old English Cairo data are also supplied for easy comparison.

\begin{enumerate}
\small
\item \textit{Sēo mǣden wrāt ǣnne ærendgecwidd tō hyre frēonde.}

``The girl wrote a letter to her friend.''

Þæt mæden ƿrat hyre freonde ærend-ƿrit.

\item \textit{Ic þence þæt hit rīne.}

``I think that it is raining.''

Ic þence þæt hit reȝnaþ.

\item \textit{Hē ġewende tō geswīgenne smōkunge and drincan.} 

``He tried to stop smoking and drinking.''

He ȝesohte to ȝesƿicanne smican and drincan.

\item \textit{Wilt þū gān?}

``Do you want to go?''

Ƿilt þu gan?

\item \textit{Sam, ontyne þæt ēagþȳrel!}

``Sam, open the window!''

Sam, ontyne þæt eagþyrel!

\item \textit{Hēo dyde þæt hyre efenġemacca þæt cræt gewēasce}

``She made her husband wash the car.''

Heo dyde þæt hyre efen-ȝemacca þæt cræt ȝeƿeasce.

\item \textit{Pēteres nēahġebūr smearde þone stoccweall rēad.}

``Peter’s neighbor painted the fence red.''

Peteres neah-ȝebur hæfð þone eodor readne onȝemet.

\item \textit{Mīn fæder is cealdra þonne þīn.}

``My dad is cooler than yours.''

Min fæder is micelre þonne þin.

\item \textit{Maria gewann brons, Pēter seolfor, and Ġeane gold.}

``Mary won bronze, Peter silver, and Jane gold.''

Maria ȝeƿann ær, Petrus seolfor, and Iohanna gold.

\item \textit{Is Iguazu micel ríce oþþe lȳtel?}

``Is Iguazu a big or a small country?''

Is Iguazu micel rice oþþe lytel?

\item \textit{Nǣþer Pēter Smīþ nē Maria Brūn mihton bēon gecōrene.} 

``Neither Peter Smith nor Mary Brown could be selected.''

Ne Petrus Smiþ ne Maria Brun mihton beon ȝecorene.

\item \textit{Hīe nabbaþ nāne ġeþōht hwā hit wrāt.} 

``They have no idea who wrote it.''

Hie nabbaþ nane ȝeþoht hƿa hit ƿrat.

\item \textit{Hwæt locast þū on?} 

``What are you looking at?''

Hƿæt locast þu on?

\item \textit{Hwonne þenctst þū þæt þū mōst cuman?} 

``When do you think you can come?''

Hƿonne þenctst þu þæt þu meaht cuman?

\item \textit{Hē bohte cræt ac his brōþor hweól-bearwe.} 

``He bought a car but his brother just a wheelbarrow.''

He bohte cræt ac his broþor hƿeol-bearƿe.

\item \textit{Petrus and Maria beclypton heom be-tweonen and þā forlētone þā heoþe.} 

``Peter and Mary hugged each other and then left the room.''

Petrus and Maria beclypton heom be-tƿeonen and þa forletone þa heoþe.

\item \textit{Hēo scolde þæt hire hēafodġearwung dyde ac for sumre rǣswan hēo nolde on þām dæġe.} 

``She should have been doing her hair but for some reason she wouldn't that day.''

Hit ȝebyrede þt heo dyde hire her, ac for sume ƿisan nolde seo don sƿa.

\item \textit{Ic ne mihte fylġan, forþām hē tō hrædlīce arn.} 

``I wasn't able to keep up, because he ran too fast.''

Ic ne mihte þurhƿunian, forðon he to sƿiftlice rann.

\item \textit{Þis gewrit is fram Pētere and hit wæs ġebrōht ġierstan dæġe.} 

``This letter is from Peter and it was delivered yesterday.''

Þis ærend-ƿrit is fram Petere and hit ƿæs ȝebroht ȝierstan-dæg.

\item \textit{Hēo weox on Paris, þǣre heofodburh Franclandes.} 

``She grew up in Paris, the capital of France.''

Heo ƿeox on Paris, þære heafod-burh Franc-landes.
\end{enumerate}

\section{Sample OE sentence analysis}
\label{appendix:sample}
We provide a sample Old English sentence analysis, complete with its syntactic features, hyperlemma, Indo-European root, and morphological features in Figure \ref{fig:sample}.

\section{Old English specific morphological feature space}
\label{appendix:feats}

Our feature space follows that of Modern English, with the exception of case and gender, with 5 cases (nominative, accusative, genitive, dative, instrumental -- the latter not attested in UD Cairo) and 3 genders (masculine, neuter, and feminine). Old English also exhibited dual number in pronouns, though this is not attested in our data. Features in tense, person, verb-form, and mood are equivalent with constraints set in Modern English UD corpora.

\begin{enumerate}
\small
    \item Case. \texttt{Nom, Acc, Gen, Dat, Ins}.
    \item Gender. \texttt{Masc, Fem, Neut}.
    \item Number. \texttt{Sing, Dual, Plur}.
    \item Person. \texttt{1, 2, 3}.
    \item VerbForm. \texttt{Fin, Inf, Part, Ger}.
    \item Mood. \texttt{Ind, Imp, Sub}.
    \item Tense. \texttt{Pres, Pret}
    \item Degree. \texttt{Pos, Comp, Sup}
    \item Poss. \texttt{Yes}
\end{enumerate}

\end{document}